\documentclass[11pt,a4paper]{article}
\usepackage[hyperref]{acl2020}
\usepackage{times}
\usepackage{latexsym}

\usepackage{microtype}
\usepackage[english]{babel}
\usepackage[utf8]{inputenc}
\usepackage{multicol}
\usepackage{multirow}
\usepackage{algorithm}
\usepackage{amsmath}
\usepackage{graphicx}
\usepackage{amssymb}
\aclfinalcopy

\title{Event Extraction: A Survey}

\author{Viet Dac Lai\\
Department of Computer Science \\
University of Oregon\\
Eugene, OR 97403, USA \\
\textit{vietl@cs.uoregon.edu}}

\date{}

\begin{document}
\maketitle
\begin{abstract}
Extracting the reported events from text is one of the key research themes in natural language processing. This process includes several tasks such as event detection, argument extraction, role labeling. As one of the most important topics in natural language processing and natural language understanding, the applications of event extraction spans across a wide range of domains such as newswire, biomedical domain, history and humanity, and cyber security. This report presents a comprehensive survey for event detection from textual documents. In this report, we provide the task definition, the evaluation method, as well as the benchmark datasets and a taxonomy of methodologies for event extraction. We also present our vision of future research direction in event detection.

\end{abstract}

\newcommand{\tablett}[1]{\multicolumn{1}{|c|}{\textbf{#1}}}
\newcommand{\ttmc}[2]{\multicolumn{#1}{|c|}{\textbf{#2}}}
\newcommand{\ttmr}[2]{\multirow{#1}{*}{\textbf{#2}}}
\newcommand{\ttrt}[1]{\rotatebox{90}{\textbf{#1} }}
\newcommand{\tk}[0]{$\checkmark$}

\section{Introduction}

Event Extraction (EE) is an essential task in Information Extraction (IE) in Natural Language Processing (NLP). An event is an occurrence of an activity that happens at a particular time and place, or it might be described as a change of state \cite{ldc2005ace}. The main task of event extraction is to detect events in the text (i.e., event detection) and then sort them into some classes of interest (i.e., event classification). The second task involves detecting the event participants (i.e., argument extraction) and their attributes (e.g., argument role labeling). In short, event extraction structures the unstructured text by answering the \textit{WH} questions of an event (i.e., what, who, when, where, why, and how).

Event extraction plays an important role in various natural language processing applications. For instance, the extracted event can be used to construct knowledge bases on which people can perform logical queries easily \cite{ge2018eventwiki}. Many domains can benefit from the development of event extraction research. In the biomedical domain, event extraction can be used to extract interaction between biomolecules (e.g., protein-protein interactions) that have been described in the biomedical literature \cite{kim2009overview}. In the economic domain, events reported on social media and social networks can be used for measuring socio-economic indicators \cite{min2019measure}. Recently, event extraction has been adopted in many other domains such as: literature \cite{sims-etal-2019-literary}, cyber security \cite{man-duc-trong-etal-2020-introducing}, history \cite{sprugnoli-tonelli-2019-novel}, and humanity \cite{lai-etal-2021-event}.

Even though event extraction has been studied for decades, it is still a very challenging task. To perform the event extraction, a system needs to understand the text's semantics and ambiguity and organize the extracted information into structures \cite{ldc2005ace}.

It closely connects with other natural language processing tasks such as named entity recognition (NER), entity linking (EL), and dependency parsing. Although these tasks can boost the development of event extraction \cite{mcclosky2011event}, they might have an inverse impact on the performance of the event extraction systems \cite{zhang2018graph}, depending on how the output of these tasks is exploited.

Last but not least, lacking training data is a fundamental problem in expanding event extraction to a new domain because the traditional classification model requires a large amount of training data \cite{huang:18:zeroshot}. Therefore, extracting events with a substantially small amount of training data is a new and challenging problem.
\section{The event extraction task}

\subsection{Definition}

Event extraction aims to detect the appearance of an event in the text (e.g., sentence, document) and its related information. In some cases, a predefined structure of the event is provided to formulate the event, such as participants and their relations to the event. This is called close-domain event extraction, as the expected structure is provided according to a certain application. Otherwise, open-domain event extraction does not require any predefined structure. In this report, we focus on close-domain event extraction.

ACE-2005 \cite{ldc2005ace} defines an event schema whose terminologies have been widely used in event extraction:

\begin{itemize}
    \item Event extent is a sentence within which an event is expressed.
    \item Event trigger is a word or a phrase that most clearly expresses the occurrence of the event. In many cases, the event trigger is the main verb of the sentence expressing the event.
    \item Event arguments are entities that are part of the event. They include participants and attributes.
    \item Argument role is the relationship between an event and its arguments.
\end{itemize}

Based on these terminologies, \citet{Ahn:06} proposes to divide the event extraction into sub-tasks: trigger detection, trigger classification, argument detection, and argument classification. 

\textit{Earlier documents in the case have included embarrassing details about perks \textbf{Welch} received as part of \textbf{his} \underline{retirement} package from \textbf{GE} at a time when corporate scandals were sparking outrage.}

\begin{table}[!h]
    \centering
    \begin{tabular}{|l|c|}
    \hline
       Trigger   & retirement \\
       \hline
       Event type  & Personnel:End-Position\\
       \hline
       Person-Arg & Welch\\
       Entity-Arg & GE \\
       Position-Arg & - \\
       Time-Arg & -\\
       Place-Arg & -\\
       \hline
    \end{tabular}
    \caption{Example of a sample in ACE-2005.}
    \label{tab:example-ace}
\end{table}

\subsection{Corpora}
\label{sec:corpora}

The development of event extraction is mostly promoted by the availability of data offered by public evaluation programs such as Message Understanding Conference (MUC), Automatic Content Extraction (ACE), and Knowledge Base Population (TAC-KBP).

\textbf{Automatic Content Extraction (ACE-2005)} is the most widely used corpus in event extraction for English, Arabic, and Chinese. It annotates entities,  events, relation, and time \cite{ldc2005ace}. There are 7 categories of entities in ACE-2005 i.e., person, organization, location, geopolitical entity, facility, vehicle, and weapon. The ACE-2005 defines 8 event types and 33 event subtypes as presented in table \ref{tab:ace}. This dataset annotates 599 documents from various sources e.g., weblogs, broadcast news, newsgroups, and broadcast conversation.

\begin{table}[]
    \centering
    \begin{tabular}{|l|p{5cm}|}
        \hline
       \tablett{Event type}  & \tablett{Event subtype} \\
       \hline
       Life  & Be-born, Marry, Divorce, Injure, Die \\
       \hline
       Movement & Transport \\
       \hline
       Transaction & Transfer-Ownership, Transfer-Money \\
       \hline
       Business & Start-Org, Merge-Org, Declare-Bankruptcy,  End-Org \\
       \hline
       Conflict & Attack, Demonstrate \\
       \hline
       Contact & Meet, Phone-Write \\
       \hline
       Personnel & Start-Position, End-Position, Nominate, Elect \\
       \hline
       Justice & Arrest-Jail, Release-Parole, Trial-Hearing, Charge-Indict, Sue, Convict, Sentence, Fine, Execute, Extradite, Acquit, Appeal, Pardon \\
       \hline
    \end{tabular}
    \caption{List of event types and event subtypes covered in ACE-2005.}
    \label{tab:ace}
\end{table}

\textbf{TAC-KBP} datasets aim to promote extracting information from unstructured text that fits the knowledge base. The dataset includes the annotation for event detection, event coreference, event linking, argument extraction, and argument linking \cite{ellis2015overview}. The event taxonomy in TAC-KBP is mostly derived from ACE-2005, with 9 event types and 38 event subtypes. This dataset is annotated from 360 documents, of which 158 documents are used for training and 202 documents are used for testing. The TAC-KBP 2015 contains documents for English only \cite{ellis2015overview}, whereas TAC-KBP 2016 includes Chinese and Spanish documents \cite{ji2016overview}.

Many corpora for specific domains have been published for public use. \textbf{MUC} corpora annotate events for various domains such as fleet operation, terrorism, and semiconductor production \cite{grishman1996message}. The \textbf{GENIA} is an event detection corpus for the biomedical domain. It is compiled from scientific documents from PubMed by the BioNLP Shared Task \cite{kim2009overview}. \textbf{TimeBank} annotates 183 English news articles with event, temporal annotations, and their links \cite{pustejovsky2003timebank}. Recently, event detection has expands to many other fields such as \textbf{CASIE} and \textbf{CyberED} for cyber-security \cite{satyapanich2020casie,trong2020introducing}, \textbf{Litbank} for literature \cite{sims-etal-2019-literary}, and music \cite{ding2011research}. However, these corpora are both small in the number of data samples and close in terms of the domain. Consequently, this limits the ability of the pre-trained models to perform tasks in a new domain in real applications.

On the other hand, a general-domain dataset for event detection is a good fit for real applications because it offers a much more comprehensive range of domains and topics. However, manually creating a large-scale general-domain dataset for ED is too costly to anyone ever attempt. Instead, general-domain datasets for event detection have been produced at a large scale by exploiting a knowledge base and unlabeled text. Distant supervision and learning models are two main methods that have been employed to generate large-scale ED datasets.

Distant supervision \cite{mintz-etal-2009-distant} is the most widely use with facts derived from existing knowledge base such as WordNet \cite{miller1995wordnet}, FrameNet \cite{baker1998berkeley}, and Freebase \cite{bollacker2008freebase}.

\citet{chen-etal-2017-automatically} proposes an approach to align key arguments of an event by using Freebase. Then these arguments are used to detect the event and its trigger word automatically. The data is further denoised by using FrameNet \cite{baker1998berkeley}. Similarly, \cite{wang-etal-2020-maven} constructs the \textbf{MAVEN} dataset from Wikipedia text and FrameNet. This dataset also offers a tree-like event schema structure rooted in the word sense hierarchy in FrameNet. 

Similarly, \citet{le-nguyen-2021-fine} creates \textbf{FedSemcor} from WordNet and Word Sense Disambiguation dataset. A subset of WordNet synsets that are more likely eventive is collected and grouped into event detection classes with similar meanings. The Semcor is a word sense disambiguation dataset whose tokens are labeled by WordNet synsets. As such, to create the event detection, the text from the Semcor dataset is realigned with the collected ED classes.

Table \ref{tab:dataset_statistics} presents a summary of the existing event extraction dataset for English. 


\begin{table*}[t]
\centering
\resizebox{\textwidth}{!}{
\begin{tabular}{|l|l|r|r|c|l|}
    \hline
    \tablett{Dataset} & \tablett{Topic} & \ttrt{\#Class} & \ttrt{\#Samples} & \ttrt{\#Language} &  \tablett{Tasks}\\
    \hline
    \ttmc{6}{Event Extraction} \\
    \hline
    ACE             & News              & 33    & 4,907     & 3     & Trig, Arg, Ent, Rel,\\
    \cite{ldc2005ace} &&& events &                                   & Coreference \\ \hline 
    TAC-KBP         & News              & 38    & 11,975    & 3     & Trig, Arg, Ent, Rel, \\
    \cite{ellis2015overview} &&& events &                                          & Coreference \\ \hline 
    TimeBank        & Newswire          & 8     & 7,935     & 1     & Trig \\ 
    \cite{pustejovsky2003timebank} &&& events && Temporal \\ 
    \hline 
    GENIA           & Biomedical        & 36    & 36,114    & 1     & Trig, Arg, Ent, Rel \\
    \cite{ohta2002genia} &&                     & events & &  \\ \hline 
    CASIE           & Cyber security    & 5     & 8,470     & 1     & Trig \\
    \cite{satyapanich2020casie} &&& events & &  \\ \hline 
    CyberED        & Cyber security    & 30    & 8,014     & 1     & Trig \\ 
    \cite{man-duc-trong-etal-2020-introducing} &&& events &&  \\ \hline 
    Litbank \cite{sims-etal-2019-literary}        & Literature        & 1     & 7,849     & 1     & Trig, Ent \\
    \cite{bamman-etal-2019-annotated,bamman-etal-2020-annotated} &&& events &                                                     & EntCoref \\ \hline 
    TranscriptED    & Graphical design  &       &           &       & Trig \\ 
    \cite{} &&& events && \\ \hline 
    BRAD            & Black rebellion   & 12    & 4,259     & 1     & Trig, Arg, Ent, Rel \\ \cite{lai-etal-2021-event} &&& events &&  \\ \hline 
    SuicideED       & Mental health     & 7     & 36,978    & 1     & Trig, Arg, Ent, Rel \\
    \cite{guzman-nateras-etal-2022-event} &&& events && \\ \hline 
    MAVEN           & General           & 168   &  111,611  & 1     & Trig \\ 
    \cite{wang-etal-2020-maven} &&& events &                                                     &  \\ \hline 
    FedSemcor       & General           & 449   & 34,666    & 1     & Trig \\ 
    \cite{le-nguyen-2021-fine} &&& events & & \\ \hline
    MINION          & Economy, Politics,    & &&&\\ 
    \cite{pouran-ben-veyseh-etal-2022-minion}  & Technology, Nature,   & 33    & 50,934    & 10    & Trig \\ 
            & Crime, Military       & &&&\\ \hline
    CLIP-Event  & News              &   33    & 105,331   &   1    & Trig, Arg, Ent      \\
    \cite{li2022clip} &&& events & & \\ \hline
    \ttmc{6}{Event Relation} \\
    \hline
    
    HiEve           & News stories  & 2     & 2,257     & 1 & Hierarchy,  \\  
    \cite{glavas-etal-2014-hieve}           &&& pairs       &&  Coreference \\ 
    \hline 
    TempEval        &   &&&                         & Temporal  \\
    \cite{uzzaman-etal-2013-semeval} &&& pairs &     &  \\ 
    \hline 
    EventStoryLine  & Calamity events &2    & 8,201     & 1     & Causal, \\  
    \cite{caselli-vossen-2017-event}        &&& pairs &&  Temporal\\ 
    \hline 
    
    MATRES          &                   &&&         1                 &  Temporal \\
    \cite{ning-etal-2018-multi}&&&&                 &  \\ 
    \hline 
    MECI            & Wikipedia     & 2     &  11,055   & 5     &  Causal\\ 
    \cite{lai-etal-2022-meci}               &&& pairs           &&  \\ 
    \hline 
    mSubEvent       & Wikipedia     & 2     & 3,944 & 5           &  Hierarchy\\ 
    \cite{lai-etal-2022-multilingual}       &&& pairs &&  \\ 
    \hline 
\end{tabular}
}
\caption{Statistics of existing event extraction datasets. Event-related tasks: Trigger Identificaion \& Classification (Trig),  Event Argument Extraction (Arg), Event Temporal (Temporal), Event Causality (Causal), Event Coreference (Coreference), Event Hierarchy (Hierarchy). Entity-related tasks: Entity Mention (Ent), Entity Linking (Rel), Entity Coreference (EntCoref).}
\label{tab:dataset_statistics}
\end{table*}

\section{Supervised Learning Models}

\subsection{Feature-based models}

In the early stage of event extraction, most methods utilize a large set of features (i.e., feature engineering) for statistical classifiers. The features can be derived from constituent parser \cite{Ahn:06}, dependency parser\cite{Ahn:06}, POS taggers, unsupervised topic features \cite{Liao:11}, and contextual features \cite{Patwardhan:09}. These models employ statistical models such as nearest neighbor \cite{Ahn:06}, maximum-entropy classifier \cite{Liao:11}, and conditional random field \cite{majumder2015event}.

\citet{Ahn:06} employed a rich feature set of lexical features, dependency features, and entity features. The lexical features consist of the word and its lemma, lowercase, and Part-of-Speech (POS) tag. The dependency features include the depth of the word in the dependency tree, the dependency relation of the trigger, and the POS of the connected nodes. The context features consist of left/right context, such as lowercase, POS tag, and entity type. The entity features include the number of dependants, labels, constituent headwords, the number of entities along a dependency path, and the length of the path to the closest entity.

\citet{Ji:08} further introduced cross-sentence and cross-document rules to mandate the consistencies of the classification of triggers and their arguments in a document. In particular, they include (1) the consistency of word sense across sentences in related documents and (2) the consistency of roles and entity types for different mentions of the related events.

\citet{Patwardhan:09} suggest using contextual features such as the lexical head of the candidate, the semantic class of the lexical head, lexico-semantic pattern surrounding the candidate. This information provides rich contextual features of the words surrounding the candidate and its lexical-connected words, which provides some signal for the success of convolutional neural networks and graph convolutional neural networks based on the dependency graph in recent studies.

\citet{Liao:11} shows that global topic features can help improve EE performance on test data, especially for a balanced corpus. The unsupervised topic model trained on large untagged corpus can provide underlying relations between event and entity types. Therefore, it can reduce the bias introduced in an imbalanced corpus (e.g., ACE-2005 dataset).

\citet{majumder2015event} extracts various features for biomedical event extraction, such as dependency path and distance to the nearest protein entity. Since the terminologies in the biomedical domain follow some particular rules, the suffix-prefix of words provides substantial semantic information about the terms.

Even though tremendous effort has been poured into feature engineering, feature-based models with statistical classifiers hinder the application of event extraction models in practical situations for two reasons. The first reason is the need for the manual design of the feature set, which requires research expertise in both linguistics and the target-specific domain. Second, since feature extraction tools are imperfect, their incorrect extracted features can harm the statistical models. Hence, a model which can automatically learn would significantly boost the application of event extraction.

\subsection{Neural-based models}

As mentioned in the previous section, crafting a diverse set of lexical, syntactic, semantic, and topic features requires both linguistic and domain expertise. This might hinder the adaptability of the model to real applications where expertise is scarce. Therefore, instead of manually designing linguistic features, automatically extracting features is more practical in virtually every NLP task. Hence, it can revolutionize the common practice of NLP studies. Toward this end, the deep neural network is the perfect match because of its ability to capture features from text automatically.

Deep neural networks employing multiple layers of a large number of artificial neurons have been adapted to various classification and generation tasks. In an artificial neural network, a layer takes input from the output of the lower layer and transforms it into a more abstract representation with two exceptions. The lowest layer takes input as a vector generated from the data sample. The highest layer usually outputs a score for each of the classification classes. These scores are used for the prediction of the label.


\subsubsection{Distributed word embedding}

Distributed word embedding is one of the most impactful tools for most NLP tasks, including event extraction. Word embedding plays a vital role in transitioning from feature-based to neural-based modeling. The representation obtained from word embedding captures a rich set of syntactic features, semantic features, and knowledge learned from a large amount of text \cite{mikolov2013distributed}. 

Technically, distributed word embedding is a matrix that can be viewed as a list of low-dimensional continuous float vectors \cite{bengio:03}. Word embedding maps a word into a single vector within its dictionary. Hence, a sentence can be encoded into a list of vectors. These vectors are fed into the neural network. Among tens of variants, Word2Vec \cite{mikolov2013distributed} and GloVe \cite{pennington-etal-2014-glove} are the most popular word embeddings. These word embeddings were then called context-free embedding to distinguish against contextualized word embedding, which was invented a few years after context-free word embedding.

\begin{figure*}[t]
    \centering
    \includegraphics[width=\textwidth]{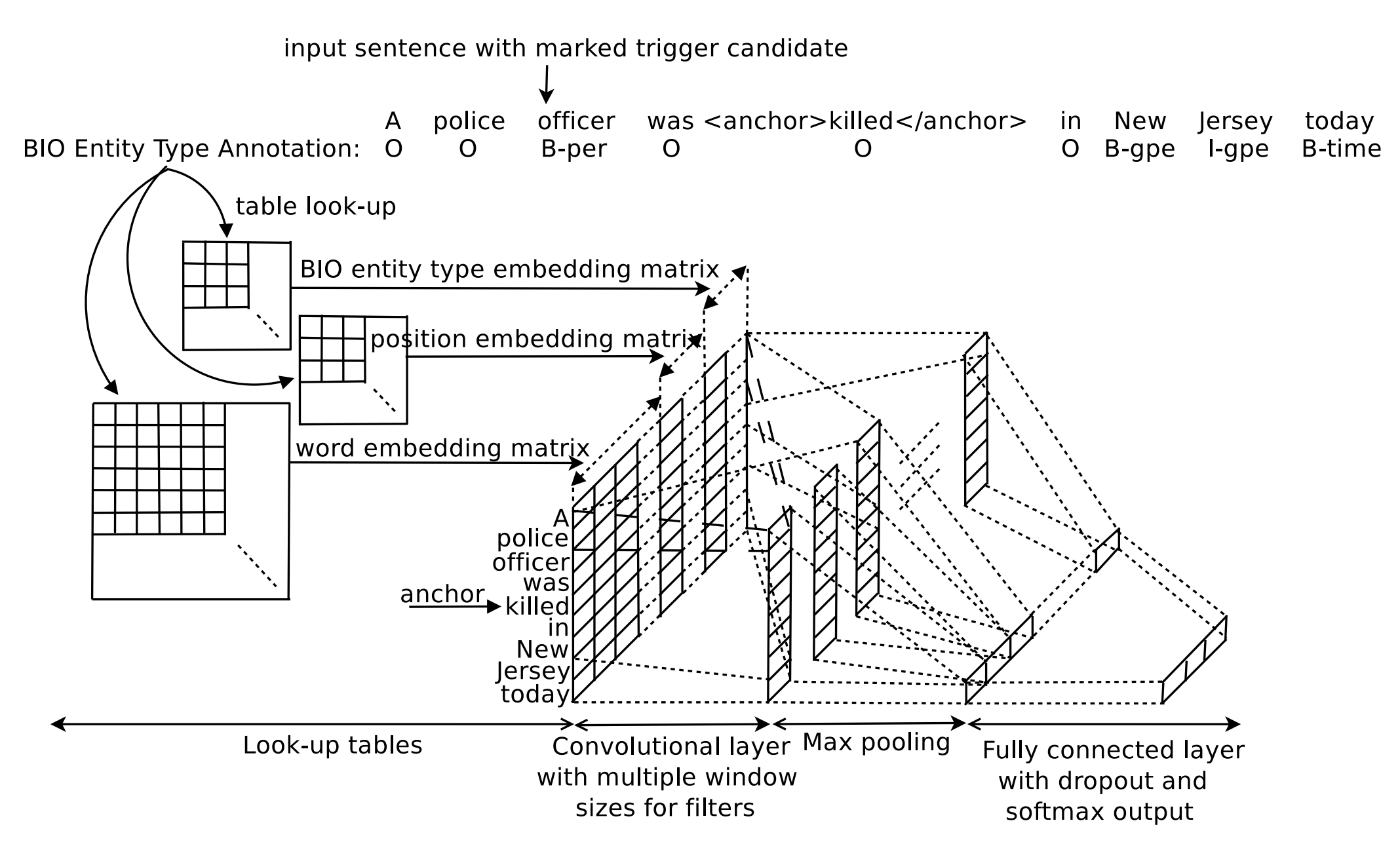}
    \caption{The convolutional neural network model for event detection in \cite{nguyen-grishman-2015-event}.}
    \label{fig:cnn-nguyen-15}
\end{figure*}

Contextualized word embedding is one of the greatest inventions in the field of NLP recently. Contrary to context-free word embedding, contextualized embedding dynamically encodes the word in a sentence based on the context presented in the text \cite{peters-etal-2018-deep}. In addition, the contextualized embeddings are usually trained on a large text corpus. Hence, its embedding encodes a substantial amount of knowledge from the text. These lead to the improvement of virtually every model in NLP. There have been many variants of contextualized word embedding for general English text e.g. BERT \cite{devlin-etal-2019-bert}, RoBERTa \cite{liu2019roberta}, multi-lingual text e.g. mBERT \cite{devlin-etal-2019-bert}, XLM-RoBERTa \cite{ruder-etal-2019-unsupervised}, scientific document SciBERT \cite{beltagy-etal-2019-scibert}, text generation e.g. GPT2 \cite{radford2019language}.


\subsubsection{Convolutional Neural Networks}

\citet{nguyen-grishman-2015-event} employed a convolutional neural network, inspired by CNNs in computer vision \cite{lecun1998gradient} and NLP \cite{kalchbrenner-etal-2014-convolutional}, that automatically learns the features from the text, and minimizes the effort spent on feature extraction. Instead of producing a large vector representation for each sample, i.e., tens of thousands of dimensions, this model employs three much smaller word embedding vectors with just a few hundred dimensions. As shown in Figure \ref{fig:cnn-nguyen-15}, given a sentence with marked entities, each word in the sentence is represented by a low-dimension vector concatenated from (1)the word embedding, (2) the relative position embedding, and (3) the entity type embedding. The vectors of words then form a matrix working as the representation of the sentence. The matrix is then fed to multiple stacks of a convolutional layer, a max-pooling layer, and a fully connected layer. The model is trained using the gradient descent algorithm with cross-entropy loss. Some regularization techniques are applied to improve the model, such as mini-batch training, adaptive learning rate optimizer, and weight normalization.

Many efforts have introduced different pooling techniques to extract meaningful information for event extract from what is provided in the sentence. \citet{chen-etal-2015-event} improved the CNN model by using multi-pooling (DMCNN) instead of vanilla max-pooling. In this model, the sentence is split into multiple parts by either the examining event trigger or the given entity markers. The pooling layer is applied separately on each part of the sentence. \citet{zhang2016joint} proposed skip-window convolution neural networks (S-CNNs) to extract global structured features. The model effectively captures the global dependencies of every token in the sentence. \citet{li2018extracting} proposed a parallel multi-pooling convolutional neural network (PMCNN) that applies not only multiple pooling for the examining event trigger and entities but also to every other trigger and argument that appear in the sentence. This helps to capture the compositional semantic features of the sentence.

\citet{kodelja2019exploiting} integrated the global representation of contexts beyond the sentence level into the convolutional neural network. To generate the global representation in connection with the target event detection task, they label the whole given document using a bootstrapping model. The bootstrapping model is based on the usual CNN model. The predictions for every token are aggregated to generate the global representation.

\begin{figure*}[t]
    \centering
    \includegraphics[width=\textwidth]{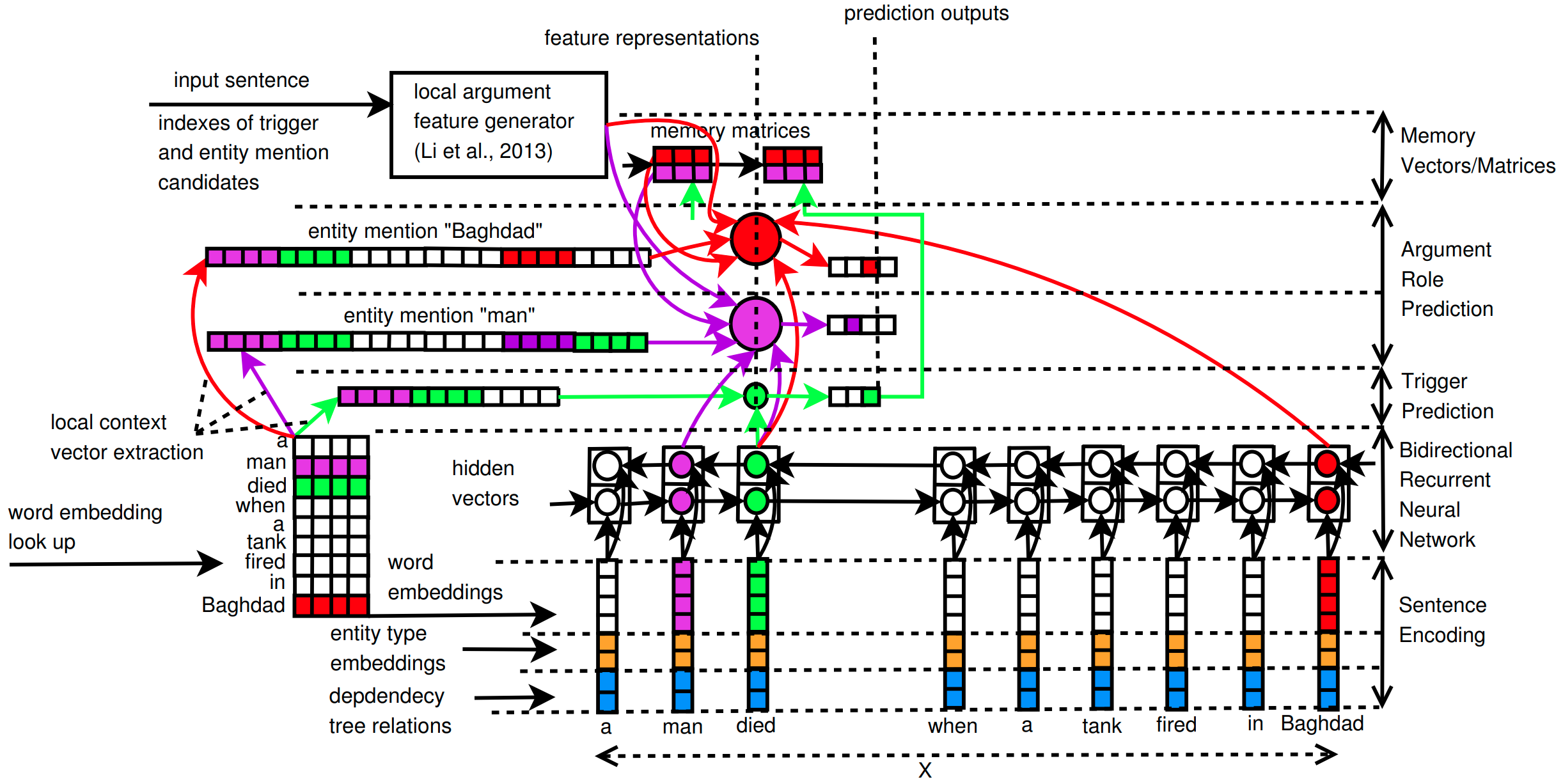}
    \caption{The joint EE model \cite{nguyen-etal-2016-joint} for the input sentence “a man died when a tank fired in Baghdad” with local context window d = 1.}
    \label{fig:rnn-nguyen-16}
\end{figure*}

Even though CNN, together with the distributed word representations, can automatically capture local features, EE models based on CNN are not successful at capturing long-range dependency between words. The reason is that CNN can only model the short-range dependencies within the window of its kernel. Moreover, a large amount of information is lost because of the pooling operations (e.g. max pooling). As such, a more sophisticated neural network design is needed to model the long-range dependency between words in long sentences and documents without sacrificing information.

\subsubsection{Recurrent Neural Networks}

\citet{nguyen-etal-2016-joint-event} first used Gated Recurrent Unit (GRU) \cite{cho-etal-2014-properties}, a RNN-based architecture, to better model relation between words in a sentence. The model produces a rich representation based on the context captured in the sentence for the prediction of event triggers and event arguments. The model includes two recurrent neural networks, one for the forward direction and one for the backward direction. A diagram of the whole model is presented in Figure \ref{fig:rnn-nguyen-16}.

\textbf{Sentence embedding}: Similar to CNN model, each word $w_i$ of the sentence is transformed into a fixed-size real-value vector $x_i$. The feature vector is a concatenation of the word embedding vector of the current word, the embedding vector for the entity type of the current word, and the one-hot vector whose dimensions correspond to the possible relations between words in the dependency trees.

\textbf{RNN encoding:} The model employs two recurrent networks , forward and backward, denoted as $\overrightarrow{RNN}$ and $\overleftarrow{RNN}$ to encode the sentence word-by-word: 
$$(a_1, \cdots, a_N )=\overrightarrow{RNN}(x_1,\cdots,x_N)$$
$$(a'_1, \cdots, a'_N ) = \overleftarrow{RNN}(x_1,\cdots,x_N)$$
Finally, the representation $h_i$ for each word is the concatenation of the corresponding forward and backward vectors $h_i=[a_i,a'_i]$.

\textbf{prediction}: To jointly predict the event triggers and arguments, a binary vector for trigger and two binary matrices are introduced for event arguments. These vectors and matrices are initialized to zero. For each iteration, according to each word $w_i$, the prediction is made in a 3-step process: trigger prediction for $w_i$, argument role prediction for all the entity mentions given in the sentence, and finally, compute the vector and matrices of the current step using the memory and the output of the previous step. 

Similarly, \citet{ghaeini-etal-2016-event} and \citet{chen2016event} employed Long Short-Term Memory (LSTM) \cite{hochreiter1997long}, anther architecture based on RNN. LSTM is much more complex than the original RNN architecture and the GRU architecture. LSTM can capture the semantics of words with consideration of the context given by the context words automatically. \citet{chen2016event} further proposed Dynamic Multi-Pooling similar to the DMCNN \cite{chen-etal-2015-event} to extract event and argument separately. Furthermore, the model proposed a tensor layer to model the interaction between candidate arguments. 

Even though the vanilla LSTM (or sequential/linear LSTM) can capture a longer dependency than CNN, in many cases, the event trigger and its arguments are distant, as such, the LSTM model can not capture the dependency between them. However, the distance between those words is much shorter in a dependency tree. Using a dependency tree to represent the relationship between words in the sentence can bring the trigger and entities close to each other. Some studies have implemented this structure in various ways. \citet{sha2018jointly} proposed to enhance the bidirectional RNN with dependency bridges, which channel the syntactic information when modeling words in the sentence. The paper illustrates that simultaneously employing hierarchical tree structure and sequence structure in RNN improves the model's performance against the conventional sequential structure. \citet{li-etal-2019-biomedical} introduced tree a  knowledge base (KB)-driven tree-structured long short-term memory networks (Tree-LSTM) framework. This model incorporates two new features: dependency structures to capture broad contexts and entity properties (types and category descriptions) from external ontologies via entity linking.  

\subsection{Graph Convolutional Neural Networks}

The presented CNN-based and LSTM-based models for event detection have only considered the sequential representation of sentences. However, in these models, graph-based representation such as syntactic dependency tree \cite{nivre-etal-2016-universal} has not been explored for event extraction, even though they provide an effective mechanism to link words to their informative context in the sentences directly. 

\begin{figure*}[t]
    \centering
    \includegraphics[width=\textwidth]{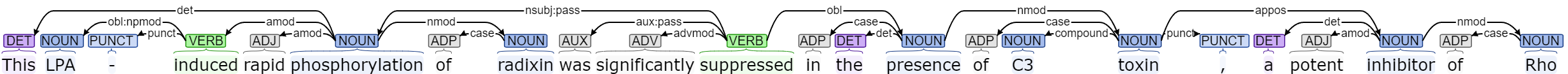}
    \caption{ Dependency tree for sentence ``\textit{This LPA-induced rapid phosphorylation of radixin was significantly suppressed in the presence of C3 toxin, a potent inhibitor of Rho}'', parsed by Trankit toolkit\footnote{http://nlp.uoregon.edu/trankit}.}
    \label{fig:dependency-tree}
\end{figure*}

For example, Figure \ref{fig:dependency-tree} presents the dependency tree of the sentence ``\textit{This LPA-induced rapid phosphorylation of radixin was significantly suppressed in the presence of C3 toxin, a potent inhibitor of Rho}''. In this sentence, there is a event trigger ``\textit{suppressed}'' with its argument ``\textit{C3 toxin}''. In the sequential representation, these words are 5-step apart, whereas in the dependency tree, they are 2-step apart. This example demonstrates the potential of the dependency tree in extracting event triggers and their arguments.

Many EE studies have widely used graph convolutional neural networks (GCN) \cite{Kipf17semisupervised}. It features two main ingredients: a convolutional operation and a graph. The convolutional operation works similarly in both CNNs and GCNs. It learns the features by integrating the features of the neighboring nodes. In GCNs, the neighborhoods are the adjacent nodes on the graph, whereas, in CNNs, the neighborhoods are surrounding words in linear form. 

Formally, let $\mathcal{G}=(\mathcal{V}, \mathcal{E})$ be a graph, and $A$ be its adjacency matrix. The output of the $l+1$ convolutional layer on a graph $\mathcal{G}$ is computed based on the hidden states $H^l=\{h^l_i\}$ of the $l$-th layer  as follows:
\begin{equation}
    h_i^{l+1} = \sigma\sum_{(i,j) \in \mathcal{E}}\alpha^l_{ij}W^lh_j^l + b^l
\label{eq:gcn}
\end{equation}
Or in matrix form:
\begin{equation}
    H^{l+1} = \sigma(\alpha^l W^lH^lA + b^l)
\end{equation}
 where $W$ and $b$ are learnable parameters and $\sigma$ is a non-linear activation function; $\alpha_{ij}$ is the weight for the edge $ij$, in the simplest way, $\alpha_{ij}=1$ for all edges.

GCN-ED \cite{nguyen2018graph} and JMEE \cite{liu-etal-2018-jointly} models are the first to use GCN for event detection. The graph used in the model is based on a transformation of the syntactic dependency tree. Let $\mathcal{G}_{\text{dep}}=(\mathcal{V}, \mathcal{E}_{\text{dep}})$ be an acyclic directed graph, representing the syntactic dependency tree of a given sentence. $\mathcal{V}=\{w_i|i\in[1,N]\}$ is the set of nodes; $\mathcal{E}_{\text{dep}}=\{(w_i,w_j)|i,j\in[1,N]\}$ is the set of edges. Each node of the graph represents a token in the given sentence, whereas each directed edge represents a syntactic arc in the dependency tree. The graph $G$ used in GCN-ED and JMEE is derived with two main improvements:
\begin{itemize}
    \item For each node $w_i$, a self-loop edge $(w_i, w_i)$ is added to the set of edges so that the representation of the node is computed of the representation of itself.
    \item For each edge $(w_i, w_j)$, a reverse edge $(w_j, w_i)$ of the same dependency type is added to the set of edges of the graph.
\end{itemize}

 \begin{figure}
    \centering
    \includegraphics[width=\linewidth]{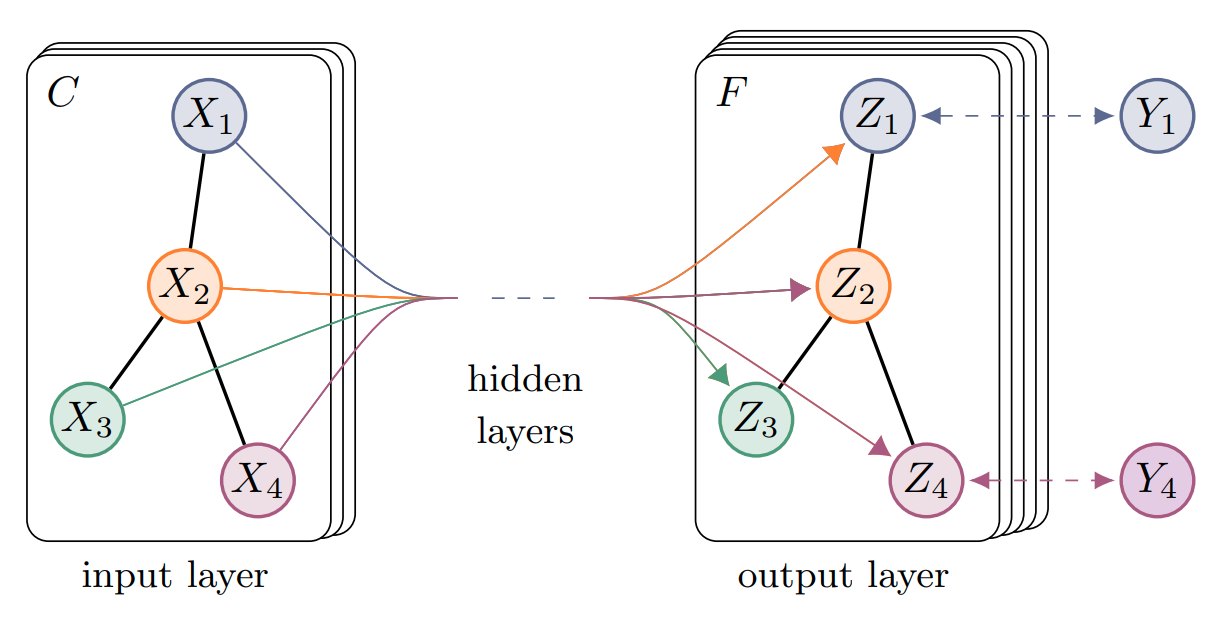}
    \caption{Graph convolutional neural network \cite{Kipf17semisupervised}.}
    \label{fig:my_label}
\end{figure}

Mathematically, a new set of edge $\mathcal{E}$ is created as follows:
\begin{equation*}
\begin{aligned}
\mathcal{E}=\mathcal{E}_{\text{dep}}&\cup \{(w_i, w_i)|w_i\in \mathcal{V}\}\\
    &\cup \{(w_j, w_i)|(w_i, w_j)\in \mathcal{E}_{\text{dep}}\}
\end{aligned}
\end{equation*}

Once the graph $\mathcal{G}=(\mathcal{V}, \mathcal{E})$ is created, the convolutional operation, as shown in Equation \ref{eq:gcn} is applied multiple times on the input word embedding. Due to the small scale of the ED dataset, instead of using different sets of weights and biases for each dependency relation type, \citet{nguyen2018graph} used only three sets of weights and biases for three types of dependency edges based on their origin: (i) The original edges from $\mathcal{E}_{\text{dep}}$, (ii) the self-loop edges, (iii) the inverse edges.

In the dependency graph, some neighbors of a node could be more important for event detection than others. Inspired by this, \citet{nguyen2018graph} and \citet{liu-etal-2018-jointly} also introduced neighbor weighting \cite{marcheggiani-titov-2017-encoding}, in which neighbors are weighted differently depending on the level of importance. The weight $\alpha$ in Equation \ref{eq:gcn} is computed as follow:

$$
\alpha^l_{ij}=\sigma(h^l_jW^l_{\text{type(i,j)}})+b^l)
$$
where $h^l_j$ is the representation of the $j$-th words at the $l$-th layer. $W^l_{\text{type(i,j)}}$ and $b^l$ are weight and bias terms, and $\sigma$ is a non-linear activation function.

However, the above dependency-tree-based methods explicitly use only first-order syntactic edges, although they may also implicitly capture high-order syntactic relations by stacking more GCN layers. As the number of GCN layers increases, the representations of neighboring words in the dependency tree will get more and more similar since they all are calculated via those of their neighbors in the dependency tree, which damages the diversity of the representations of neighboring words. As such, \citet{yan-etal-2019-event} introduced Multi-Order Graph Attention Network for Event Detection (MOGANED). In this model, the hidden vectors are computed based on the representations of not only the first-order neighbors but also higher-order neighbors in the syntactic dependency graph. To do that, they used Graph Attention network (GAT) \cite{velivckovic2017graph} and an attention aggregation mechanism to merge its multi-order representations.

In a multi-layer GCN model, each layer has its scope of neighboring. For example, the representation of a node in the first layer is computed from the representations of its first-order neighbors only, whereas one in the second layer is computed from the representations of both first-order and second-order neighbors. As such, \citet{lai-etal-2020-event} proposed GatedGCN with an enhancement to the graph convolutional neural network with layer diversity using a gating mechanism. The mechanism helps the model to distinguish the information derived from different sources, e.g., first-order neighbors and second-order neighbors. The authors also introduced importance score consistency between model-predicted importance scores and graph-based importance scores. The graph-based importance scores are computed based on the distances between nodes in the dependency graph.

The above GCN-based models usually ignore dependency label information, which conveys rich and useful linguistic knowledge for ED. Edge-Enhanced Graph Convolution Network (EE-GCN), on the other hand, simultaneously exploited syntactic structure and typed dependency label information \cite{cui-etal-2020-edge}. The model introduces a mechanism to dynamically update the representation of node-embedding and edge-embedding according to the context presented in the neighboring nodes. Similarly, \citet{dutta-etal-2021-gtn} presented the GTN-ED model that enhanced prior GCN-based models using dependency edge information. In particular, the model learns a soft selection of edge types and composite relations (e.g., multi-hop connections, called meta-paths) among the words, thus producing heterogeneous adjacency matrices.


\subsection{Knowledge Base}

As mentioned before, event extraction extract events from the text that involves some named entities such as participants, time, and location. In some domains, such as the biomedical domain, it requires a broader knowledge acquisition and a deeper understanding of the complex context to perform the event extraction task. Fortunately, a large number of those entities and events have been recorded in existing knowledge bases. Hence, these knowledge bases may provide the model with a concrete background of the domain terminologies as well as their relationship. This section presents some methods to exploit external knowledge to enhance event extraction models.

\citet{li-etal-2019-biomedical} proposed a model to construct knowledge base concept embedding to enrich the text representation for the biomedical domain. In particular, to better capture domain-specific knowledge, the model leverages the external knowledge bases (KBs) to acquire properties of all the biomedical entities. Gene Ontology is used as their external knowledge base because it provides detailed gene information, such as gene functions and relations between them as well as gene product information, e.g., related attributes, entity names, and types. Two types of information are extracted from the KB to enrich the feature of the model: (1) entity type and (2) gene function description. First, the entity type for each entity is queried, then it is injected into the model similar to \cite{nguyen-grishman-2015-event}. Second, the gene function definition, which is usually a long phrase, is passed through a language model to obtain the embedding. Finally, the embedding is concatenated to the input representation of the LSTM model.

\citet{huang-etal-2020-biomedical}, on the other hand, argues that the word embedding does not provide adequate clues for event extraction in extreme cases such as non-indicative trigger words and nested structures. For example, in the biomedical domain, many entities have hierarchical relations that might help to provide domain knowledge to the model. In particular, the Unified Medical Language System (UMLS) is the knowledge base that is used in this study. UMLS provides a large set of medical concepts, their pair-wise relations, and relation types. To incorporate the knowledge, words in the sentence are mapped to the set of concepts, if applicable. Then they are connected using the relations provided by the KB to form a semantic graph. This graph is then used in their graph neural network.

\subsection{Data generation}

As shown in Section \ref{sec:corpora}, most of the datasets for Event Extraction were created based on human annotation, which is very laborious. As such, these datasets are limited in size, as shown in Table \ref{tab:dataset_statistics}. Moreover, these datasets are usually extremely imbalanced. These issues might hinder the learning process of the deep neural network. To solve this problem, many methods of data generation have been introduced to enlarge the EE datasets, which results in significant improvement in the performance of the EE model. 

External knowledge bases such as Freebase, Wikipedia, and FrameNet are commonly used in event generation. \citet{liu-etal-2016-leveraging} trained an ED model on the ACE dataset to predict the event label on FrameNet text to produce a semi-supervised dataset. The generated data was then further filtered using a set of global constraints based on the original annotated frame from FrameNet. \citet{huang-etal-2016-liberal}, on the other hand, employs a word-sense disambiguation model to predict the word-sense label for unlabeled text. Words that belong to a subset of verb and noun senses are considered as trigger words. To identify the event arguments for the triggers, the text is parsed into an AMR graph that provides arguments for trigger candidates. The argument role is manually mapped from AMR argument types. \citet{chen-etal-2017-automatically,zeng2018scale} proposed to automatically label training data for event extraction based on distant supervision via Freebase, Wikipedia, and FrameNet data. The Freebase provides a set of key arguments for each event type. After that, candidate sentences are searched among Wikipedia text for the appearances of key arguments. Given the sentence, the trigger word is identified by a strong 
heuristic rule.

\citet{ferguson-etal-2018-semi} proposed to use bootstrapping for event extraction. The core idea is based on the occurrence of multiple mentions of the same event instances across newswire articles from multiple sources. Hence, if an ED model detects some event mentions at high confidence from a cluster,  the model can then acquire diverse training examples by adding the other mentions from that cluster. The authors trained an ED model based on limited available training data and then used that model for data labeling on unlabeled newswire text.

\citet{yang-etal-2019-exploring} explored the method that uses a generative model to generate more data. They generated data from the golden ACE dataset in three steps. First, the arguments in a sentence are replaced with highly similar arguments found in the golden data to create a noisy sentence. Second, a language model is used to regenerate the sentence from the noisy generated sentence to create a new smoother sentence to avoid overfitting. Finally, the candidate sentences are ranked using a perplexity score to find the best-generated sentence.

\citet{tong-etal-2020-improving} argued that open-domain trigger knowledge could alleviate the lack of data and training data imbalance in the existing EE dataset. The authors proposed a novel Enrichment Knowledge Distillation (EKD) model that can generate noisy ED data from unlabeled text. Unlike the prior methods that employed rules or constraints to filter noisy data, their model used the teacher-student model to automatically distill the training data.

\subsection{Document-level Modeling}

The methods for event extraction mentioned so far have not gone beyond the sentence level. Unfortunately, this is a systematic problem as, in reality, events and their associated arguments can be mentioned across multiple sentences in a document \cite{yang-etal-2018-dcfee}. Hence, such sentence-level event extraction methods struggle to handle documents in which events and their arguments scatter across multiple sentences and multiple mentions of such events. Document-level Event Extraction (DEE) paradigm has been investigated to address the aforementioned problem of sentence-level event extraction. Many researchers have proposed methods to model document-level relations such as entity interactions, sentence interactions \cite{huang-jia-2021-exploring-sentence,xu-etal-2021-document}, reconstruct document-level structure \cite{huang-peng-2021-document}, and model long-range dependencies while encoding a lengthy document \cite{du-cardie-2020-document}.

Initial studies for DEE did not consider modeling the document-level relation properly. \citet{yang-etal-2018-dcfee} was the first attempt to explore the DEE problem on a Chinese Financial Document corpus (ChiFinAnn) by generating weakly-supervised EE data using distant supervision. Their model performs DEE in two stages. First, a sequence tagging model extracts events at the sentence level in every sentence of the document. Second, key events are detected among extracted events, and arguments are heuristically collected from all over the document. \citet{zheng-etal-2019-doc2edag}, on the other hand, proposed an end-to-end model named Doc2EDAG. The model encodes documents using a transformer-based encoder. Instead of filling the argument table, they created an entity-based directed acyclic graph to find the argument effectively through path expansion. \citet{du-cardie-2020-document} transforms the  role filler extraction into an end-to-end neural sequence learning task. They proposed a multi-granularity reader to efficiently collect information at different levels of granularity, such as sentence and paragraph levels. Therefore, it mitigates the effect of long dependencies of scattering argument in DEE.

Some studies have attempted to exploit the relationship between entities, event mentions, and sentences of the document.
\citet{huang-jia-2021-exploring-sentence} modeled the interactions between entities and sentences within long documents. In particular, instead of constructing an isolated graph for each sentence, this paper constructs a unified unweighted graph for the whole document by exploiting the relationship between sentences. Furthermore, they proposed the sentence community that consists of sentences that are related to the arguments of the same event. The model detects multiple event mentions by detecting those sentence communities. To encourage the interaction between entities, \citet{xu-etal-2021-document} proposed a Heterogeneous Graph-based Interaction Model with a Tracker (GIT) to model the global interaction between entities in a document. The graph leverages multiple document-level relations, including sentence-sentence edges, sentence-mention edges, intra mention-mention edges, and inter mention-mention edges. \citet{huang-peng-2021-document} introduced an end-to-end model featuring a structured prediction algorithm, namely  Deep Value Networks, to efficiently model cross-event dependencies for document-level event extraction. The model jointly learns entity recognition, event co-reference, and event extraction tasks that result in a richer representation and a more robust model.

\subsection{Joint Modeling}

The above works have executed the four subtasks of event extraction in a pipeline manner where the model uses the prediction of other models to perform its task. Consequently, the errors of the upstream subtasks are propagated through the downstream subtasks in the pipeline, hence, ruining their performances. Additionally, the knowledge learned from the downstream subtasks can not influence the prediction decision of the upstream subtasks. Thus, the dependence on the tasks can not be exploited thoroughly. To address the issues of the pipeline model, joint modeling of multiple event extraction subtasks is an alternative to take advantage of the interactions between the EE subtasks. The interactions between subtasks are bidirectional. Therefore, useful information can be carried across the subtasks to alleviate error propagation. 

Joint modeling can be used to train a diverse set of subtasks. For example, \citet{lee-etal-2012-joint} trained a joint model for event co-reference resolution and entity co-reference resolution, while \citet{han-etal-2019-joint} proposed a joint model for event detection and event temporal relation extraction. In the early day, modeling event detection and argument role extraction together are very popular \cite{li-etal-2013-joint,venugopal-etal-2014-relieving, nguyen-etal-2016-joint-event}. Recent joint modeling systems have trained models with up to 4 subtasks (i.e. event detection, entity extraction, event argument extraction, and entity linking) \cite{lin-etal-2020-joint,zhang-ji-2021-abstract,nguyen-etal-2021-cross,nguyen-etal-2022-joint}. Table \ref{tab:joint-model-tasks} presents a summary of the subtasks that were used for joint modeling for EE. 

Early joint models were simultaneously trained to extract the trigger mention and the argument role \cite{li-etal-2013-joint}, \citet{li-etal-2013-joint} formulated a two-task problem as a structural learning problem. They incorporated both global features and local features into a perceptron model. The trigger mention and arguments are decoded simultaneously using a beam search decoder. Later models that are based on a neural network share a sentence encoder for all the subtasks \cite{nguyen-etal-2016-joint-event,han-etal-2019-joint,wadden-etal-2019-entity} so that the training signals of different subtasks can impact the representation induced by the sentence encoder.

Besides the shared encoders, recent models use a variety of techniques to further encourage interactions between subtasks. \citet{nguyen-etal-2016-joint-event} employed a memory matrix to memorize the dependencies between event and argument labels. These memories are then used as a new type of feature in the trigger and argument prediction. They employed three types of dependencies: (i) trigger subtype dependency, (ii) argument role dependency, and (iii) trigger-argument role dependency. These terminologies were later generalized as intra/inter-subtask dependencies \cite{lin-etal-2020-joint,nguyen-etal-2021-cross,nguyen-etal-2022-joint}. 

\begin{figure*}
    \centering
    \includegraphics[width=\textwidth]{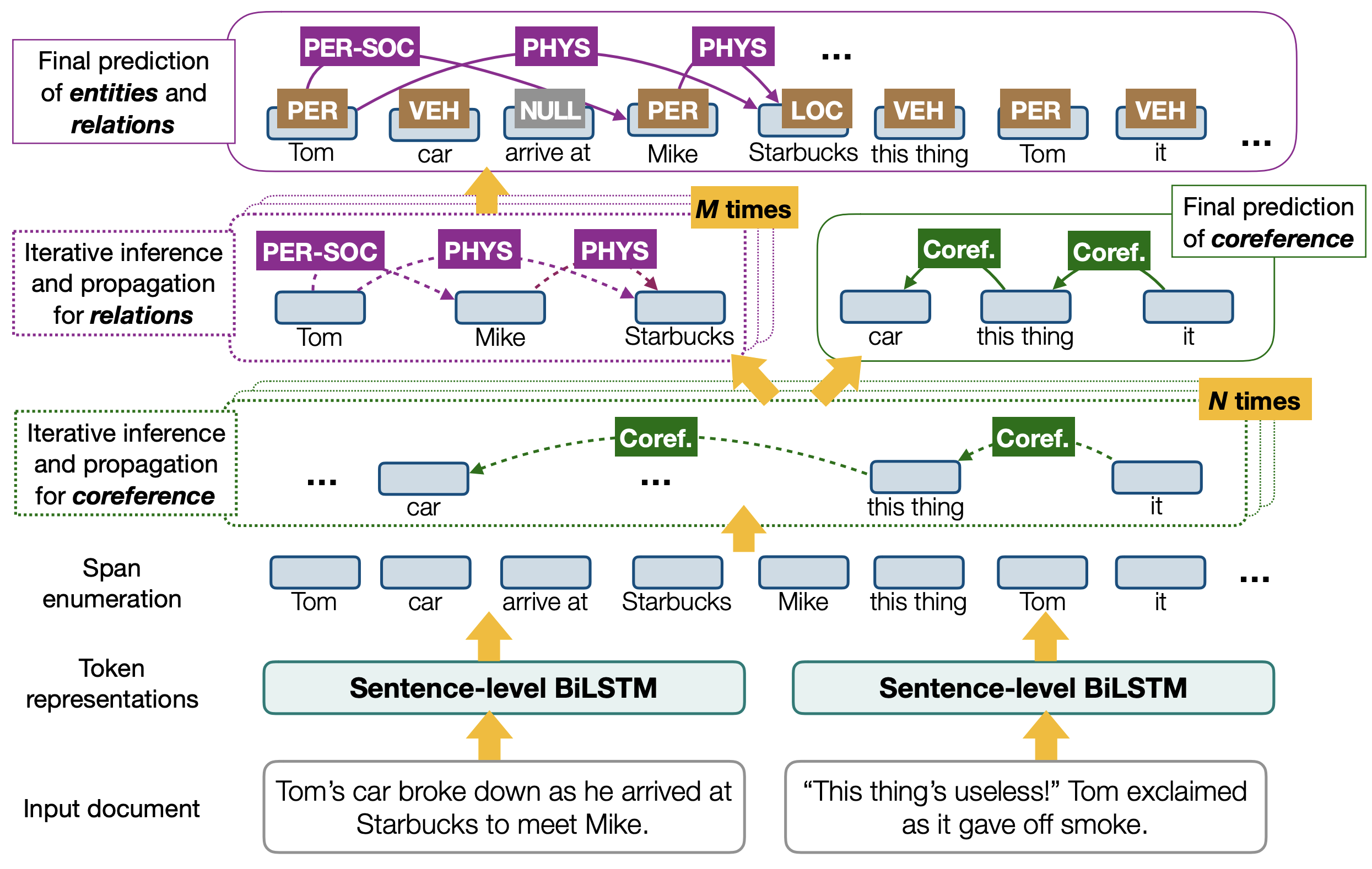}
    \caption{Architecture of the DyGIE model \cite{luan-etal-2019-general}.}
    \label{fig:luan19dygie}
\end{figure*}

\begin{table*}[t]
\centering
\begin{tabular}{|p{3cm}|p{6cm}|c|c|c|c|c|c|c|}
\hline
    \ttmr{1}{Model Acronym} & \tablett{System} & \ttrt{Event} & \ttrt{Entity} & \ttrt{Argument} & \ttrt{Relation} & \ttrt{EventCoref} & \ttrt{EntityCoref} & \ttrt{EventTemp}   \\
    \hline
    Lee's Joint & \citet{lee-etal-2012-joint}        &   &   &   &   &\tk&\tk&  \\ \hline 
    Li's Joint & \citet{li-etal-2013-joint}          &\tk&   &\tk&   &   &   & \\ \hline 
    MLN+SVM & \citet{venugopal-etal-2014-relieving}  &\tk&   &\tk&   &   &   & \\ \hline
    Araki's Joint & \citet{araki-mitamura-2015-joint}&\tk&   &   &   &\tk&   &  \\ \hline 
    JRNN    & \citet{nguyen-etal-2016-joint-event}   &\tk&\tk&\tk&   &   &   &  \\ \hline 
    Structure Joint & \citet{han-etal-2019-joint}    &\tk&   &   &   &   &   &\tk\\ \hline 
    DyGIE & \citet{luan-etal-2019-general}           &\tk&\tk&   &\tk&   &   &  \\ \hline 
    DyGIE++ & \citet{wadden-etal-2019-entity}        &\tk&\tk&   &\tk&   &   &  \\ \hline 
    HPNet   & \citet{huang-etal-2020-joint}          &\tk&   &\tk&   &   &   & \\ \hline 
    OneIE   & \citet{lin-etal-2020-joint}            &\tk&\tk&\tk&\tk&   &   &  \\ \hline 
    NGS     & \citet{wang-etal-2020-neural}          &\tk&   &\tk&   &   &   &  \\ \hline 
    Text2Event & \citet{lu-etal-2021-text2event}     &\tk&   &\tk&   &   &   &  \\ \hline 
    AMRIE & \citet{zhang-ji-2021-abstract}           &\tk&\tk&\tk&\tk&   &   &  \\ \hline 
    FourIE  & \citet{nguyen-etal-2021-cross}         &\tk&\tk&\tk&\tk&   &   &  \\ \hline 
    DEGREE  & \citet{hsu-etal-2022-degree}           &\tk&   &\tk&   &   &   &  \\ \hline 
    GraphIE & \citet{nguyen-etal-2022-joint}         &\tk&\tk&\tk&\tk&   &   &  \\ \hline 
\end{tabular}
\caption{Subtasks for joint modeling in event extraction.}
\label{tab:joint-model-tasks}
\end{table*}

\citet{luan-etal-2019-general} proposed the DyGIE model that employed an interactive graph-based propagation between events and entities nodes based on entity co-references and entity relations. In particular, in DyGIE model \cite{luan-etal-2019-general}, the input sentences are encoded using a BiLSTM model, then, a contextualized representation is computed for each possible text span. They employed a dynamic span graph whose nodes are selectively chosen from the span pool. At each training step, the model updates the set of graph nodes. It also constructs the edge weights for the newly created graph. Then, the representations of spans are updated based on neighboring entities and connected relations. Finally, the prediction of entities, events, and their relations is performed on the latest representations. Figure \ref{fig:luan19dygie} presents the diagram of the DyGIE model. \citet{wadden-etal-2019-entity} further improved the model with contextualized embeddings BERT while maintaining the core architecture of DyGIE.
Even though these models have introduced task knowledge interaction through graph propagation, their top task prediction layers still make predictions independently. In other words, the final prediction decision is still made locally. 

To address the DyGIE/DyGIE++ issue, OneIE model \cite{lin-etal-2020-joint} proposed to enforce global constraints to the final predictions. They employed a beam search decoder at the final prediction layer to globally constraint the predictions of the subtasks. Similar to JREE model \cite{nguyen-etal-2016-joint-event}, they considered both cross-subtask interactions and cross-instance interactions. To do that, they designed a set of global feature templates to capture both types of interactions. Given all the templates, the model tries to fill all possible features and learns the feature weights. During the inference, to make the final prediction, a trivial solution is an exhaustive search. However, the search space grows exponentially, leading to an infeasible problem. As such, they proposed a graph-based beam search algorithm to find the optimal graph. In each step, the beam grows with either a new node (i.e., a trigger or an entity) or a new edge (i.e., an argument role or an entity relation).

In the above neural-based models, the predictive representation of the candidates is computed independently using contextualized embedding. Consequently, the predictive representation has not considered the representations of the other related candidates. To overcome this problem, FourIE model \cite{nguyen-etal-2021-cross} features a graph structure to encourage interactions between related instances of a multi-task EE problem. \citet{nguyen-etal-2021-cross} further argued that the global feature constraint in OneIE \cite{lin-etal-2020-joint} is suboptimal because it is manually created. They instead introduced an additional graph-based neural network to score the candidate graphs. To train this scoring network, they employ Gumbel-Softmax distribution \cite{jang2017categorical} to allow gradient updates through the discrete selection process. However, due to the heuristical design of the dependency graph, the model may fail to explore other possible interactions between the instances. As such, \citet{nguyen-etal-2022-joint} explicitly model the dependencies between tasks by modeling each task instance as a node in the fully connected dependency graph. The weight for each edge is learnable, allowing a soft interaction between instances instead of hard interactions in prior works \cite{lin-etal-2020-joint,zhang-ji-2021-abstract,nguyen-etal-2021-cross}

Recently, joint modeling for event extraction was formulated as a text generation task using pre-trained generative language models such as BART \cite{lewis-etal-2020-bart}, and T5 \cite{raffel2020exploring}. In these models \cite{lu-etal-2021-text2event,hsu-etal-2022-degree}, the event mentions, entity mentions, as well as their labels and relations are generated by an attention-based autoregressive decoder. The task dependencies are encoded through the attention mechanism of the transformer-based decoder. This allows the model to flexibly learn the dependencies between tasks as well as task instances. However, to train the model, they have to assume an order of tasks and task instances that are being decoded. As a result, the model suffers from the same problem that arose in pipeline models.

\clearpage
\begin{table*}[t]
    \centering
    \begin{tabular}{|p{4cm}|p{6.5cm}|c|c|c|c|}
    \hline
      \multirow{6}{*}{\textbf{Model Acronym}} & \multirow{6}{*}{\textbf{System}} & \ttmc{2}{Trigger} & \ttmc{2}{Argument} \\
      \cline{3-6}
      & & \ttrt{Identification} & \ttrt{Classification} & \ttrt{Identification} & \ttrt{Classification}  \\
      \hline
    \multicolumn{6}{|l|}{\textbf{Feature engineering}}  \\
    \hline
                    & \citet{Ahn:06}     & 62.6 & 60.1 & 82.4 & 57.3 \\
    Cross-document  & \citet{Ji:08}      &      & 67.3 & 46.2 & 42.6 \\
    Cross-event     & \citet{Liao:10}    &   -  & 68.8 & 50.3 & 44.6 \\
    Cross-entity    & \citet{Hong:11}    &   -  & 68.3 & 53.1 & 48.3 \\
    Structure-prediction & \citet{li-etal-2013-joint} & 70.4 & 67.5 & 56.8 & 52.7 \\
    \hline
    \multicolumn{6}{|l|}{\textbf{CNN}}  \\
    \hline
    CNN         & \citet{nguyen-grishman-2015-event} & & 69.0 & - & - \\
    DMCNN       & \citet{chen-etal-2015-event} & 73.5 & 69.1 & 59.1 & 53.5 \\
    DMCNN+DS    & \citet{chen-etal-2017-automatically} & 74.3 & 70.5 & 63.3 & 55.7 \\
    \hline
    \multicolumn{6}{|l|}{\textbf{RNN}}  \\
    \hline
    JRNN & \citet{nguyen-etal-2016-joint-event} & 71.9 & 69.3 & 62.8 & 55.4 \\
    FBRNN & \citet{ghaeini-etal-2016-event} & - & 67.4 & - & - \\
    BDLSTM-TNNs & \citet{chen2016event} & 72.2 & 68.9 & 60.0 & 54.1 \\
    DLRNN & \citet{duan-etal-2017-exploiting} & - & 70.5 & - & - \\
    dbRNN & \citet{sha2018jointly} & - & 71.9  & 67.7 & 58.7 \\
    \hline
    \multicolumn{6}{|l|}{\textbf{GCN}}  \\
    \hline
    GCN-ED & \citet{nguyen2018graph} & - & 73.1 & - & - \\
    JMEE & \citet{liu-etal-2018-jointly} & 75.9 & 73.7 & 68.4 & 60.3 \\
    MOGANED & \citet{yan-etal-2019-event} & - & 75.7 & - & - \\
    MOGANED+GTN & \citet{dutta-etal-2021-gtn} & - & 76.8 & - & - \\
    GatedGCN & \citet{lai-etal-2020-event} & - & 77.6 & - & - \\
    \hline
    \multicolumn{6}{|l|}{\textbf{Data Generation \& Augmentation}}  \\
    \hline
    ANN-FN & \citet{liu-etal-2016-leveraging} & - & 70.7 & - & - \\
    Liberal & \citet{huang-etal-2016-liberal} & - & 61.8 & - & 44.8 \\
    Chen's Generation & \citet{chen-etal-2017-automatically} & 74.3 & 70.5 & 63.3 & 55.7 \\
    BLSTM-CRF-ILP$_{multi}$& \citet{zeng2018scale} & - & 82.5 & - & 37.9 \\
    EKD & \citet{tong-etal-2020-improving} & - & 78.6 & - & - \\
    GPTEDOT & \citet{pouran-ben-veyseh-etal-2021-unleash} & - & 79.2 & - & - \\
    \hline
    \multicolumn{6}{|l|}{\textbf{Document-level Modeling}}  \\
    \hline
    HBTNGMA & \citet{chen-etal-2018-collective} & - & 73.3 & - & - \\   
    DEEB-RNN & \citet{zhao-etal-2018-document} & - & 74.9 & - & - \\
    ED3C & \citet{pouran-ben-veyseh-etal-2021-modeling} & - & 79.1 & - & - \\
    \hline
    \multicolumn{6}{|l|}{\textbf{Joint Modeling}}  \\
    \hline
    DyGIE++ & \citet{wadden-etal-2019-entity} & 76.5 & 73.6 & 55.4 & 52.5 \\
    HPNet   & \citet{huang-etal-2020-joint} & 79.2 & 77.8 & 60.9& 56.8 \\
    OneIE & \citet{lin-etal-2020-joint} & - &  72.8 & - & 56.3  \\
    NGS     & \citet{wang-etal-2020-neural}  &  - & 74.6 & & 59.5 \\
    Text2event & \citet{lu-etal-2021-text2event} & - & 71.8 & - & 54.4  \\
    AMRIE & \citet{zhang-ji-2021-abstract} & - &  72.8 & - & 57.7 \\
    FourIE & \citet{nguyen-etal-2021-cross} & - &  73.3 & - & 58.3 \\
    DEGREE & \citet{hsu-etal-2022-degree} & - &  71.7 & - & 58.0 \\
    GraphIE & \citet{nguyen-etal-2022-joint} & - &  74.8 & - & 60.2  \\
    \hline
    \end{tabular}
    \caption{Performance of the presented models on ACE-05 dataset.}
    \label{tab:my_label}
\end{table*}


\section{Low-resource Event Extraction}

State-of-the-art event extraction approaches, which follow the traditional supervised learning paradigm, require great human efforts in creating high-quality annotation guidelines and annotating the data for a new event type. For each event type, language experts need to write annotation guidelines that describe the class of event and distinguish it from the other types. Then annotators are trained to label event triggers in the text to produce a large dataset. Finally, a supervised-learning-based classifier is trained on the obtained event triggers to label the target event. This labor-exhaustive process might limit the applications of event extraction in real-life scenarios. As such, approaches that require less data creation are becoming more and more attractive thanks to their fast deployment and low-cost solution. However, this line of research faces a challenging wall due to their limited access to labeled data. This section presents recent studies on low-resource event extraction in various learning paradigms and domains. The rest of the section is organized as follow: Section \ref{sec:zero-shot} highlights some methods of zero-shot learning; section \ref{sec:few-shot} presents a new clusters of recent studies in few-shot learning. Finally, methods for cross-lingual event extraction is presented in section \ref{sec:cross-lingual}.

\begin{figure*}[t]
    \centering
    \includegraphics[width=\textwidth]{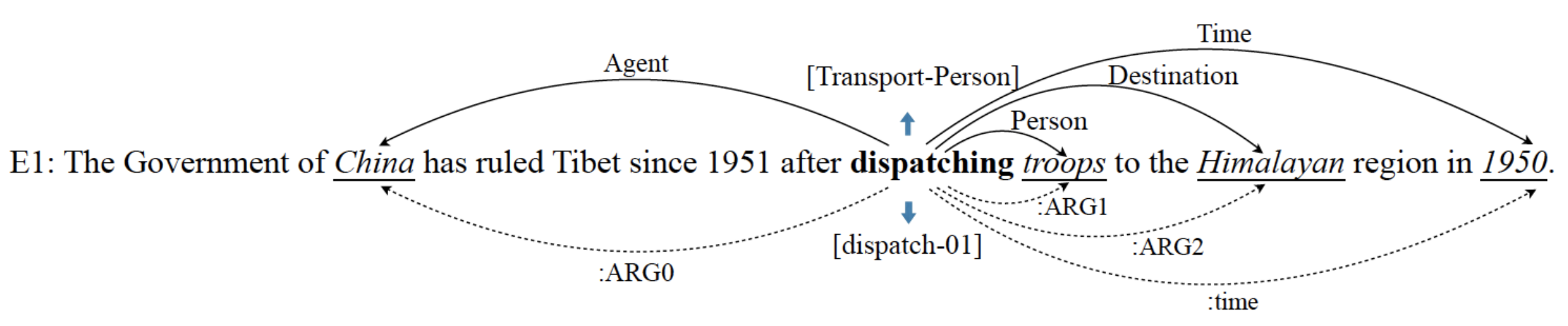}
    \caption{Comparison of AMR graph and event schema\cite{huang-etal-2018-zero}. The word \textit{dispatching} is an event trigger of the event Transport-Person with 4 arguments. The upper solid lines demonstrate the actual event arguments. The lower dash lines represent the edges in the AMR graph.}
    \label{fig:huang-18-zero-amr}
\end{figure*}

\subsection{Zero-shot Learning}

\label{sec:zero-shot}
Zero-shot learning (ZSL) is a type of transfer learning in which a model performs a task without any training samples. Toward this end, transfer learning uses a pre-existing classifier to build a universal concept space for both seen and unseen samples. Existing methods for event extraction exploits latent-variable space in CRF model \cite{lu-roth-2012-automatic}, rich structural features such as dependency tree and AMR graph \cite{huang-etal-2018-zero}, ontology mapping \cite{zhang-etal-2021-zero}, and casting the problem into a question-answering problem \cite{liu-etal-2020-event,lyu-etal-2021-zero}.

The early study by \citet{lu-roth-2012-automatic} showed the first attempt to solve the event extraction problem under zero-shot learning. They proposed to model the problem using latent variable semi-Markov conditional random fields. The model jointly extracts event mentions and event arguments given event templates, coarse event/entity mentions, and their types. They used a framework called structured Preference Modeling (PM). This framework allows arbitrary preferences associated with specific structures during the training process.

Inspired by the shared structure between events, \citet{huang-etal-2018-zero} introduced a transfer learning method that matches the structural similarity of the event in the text. They proposed a transferable architecture of structural and compositional neural networks to jointly produce to represent event mentions, their types, and their arguments in a shared latent space. This framework allows for predicting the semantically closest event types for each event mention. Hence, this framework can be applied to unseen event types by exploiting the limited manual annotations. In particular, event candidates and argument candidates are detected by exploiting the AMR graph by parsing the given text using an AMR parser. Figure \ref{fig:huang-18-zero-amr} presents an example of the AMR graph. After this, a CNN is used to encode all the triplets representing AMR edges, e.g. (dispatch-01, :ARG0, China). For each new event type, the same CNN model encodes the relations between event type, argument role, and entity type, e.g. (Transport Person, Destination), resulting in a representation vector for the new event ontology. The model chooses the closest event type based on the similarity score between the trigger's encoded vector and all available event ontology vectors to predict the event type for a candidate event trigger.

\citet{zhang-etal-2021-zero} proposed a zero-shot event extraction method that (1) extracts the event mentions using existing tools, then, and (2) maps these events to the targeted event types with zero-shot learning. Specifically, an event-type representation is induced by a large pre-trained language model using the event definition for each event type. Similarly, event mentions and entity mentions are encoded into vectors using a pre-trained language model. Initial predictions are obtained by computing the cosine similarities between label and event representations. To train the model, an ILP solver is employed to regulate the predictions according to the given ontology of each event type. In detail, they used the following constraints: (1) one event type per event mention, (2) one argument role per argument, (3) different arguments must have different types, (4) predicted triggers and argument types must be in the ontology, and (5) entity type of the argument must match the requirement in the ontology.

Thanks to the rapid development of large generative language models, a language model can embed texts and answer human-language questions in a human-friendly way using its large deep knowledge obtained from massive training data. \citet{liu-etal-2020-event} proposed a new learning setting of event extraction. They cast it as a machine reading comprehension problem (MRC). The modeling includes (1) an unsupervised question generation process, which can transfer event schema into a set of natural questions, AND (2) a BERT-based question-answering process to generate the answers as EE results. This learning paradigm exploits the learned knowledge of the language model and strengthens EE's reasoning process by integrating sophisticated MRC models into the EE model. Moreover, it can alleviate the data scarcity issue by transferring the knowledge of MRC datasets to train EE models. \citet{lyu-etal-2021-zero}, on the other hand, explore the Textual Entailment (TE) task and/or Question Answering (QA) task for zero-shot event extraction. Specifically, they cast the event trigger detection as a TE task, in which the TE model predicts the level of entailment of a hypothesis  (e.g., \textit{This is about a birth event} given a premise, i.e., the original text. Since an event may associate with multiple arguments, they cast the event argument extraction into a QA task. Given an input text and the extracted event trigger,
the model is asked a set of questions based on the event type
definition in the ontology, and retrieve the QA answers as predicted argument.

\subsection{Few-shot Learning}
\label{sec:few-shot}

There are several ways of modeling the event detection in few-shot learning scheme (FSL-ED): (1) token classification FSL-ED and (2) sequence labeling FSL-ED.

\begin{figure*}[!h]
    \centering
    \includegraphics[width=0.7\textwidth]{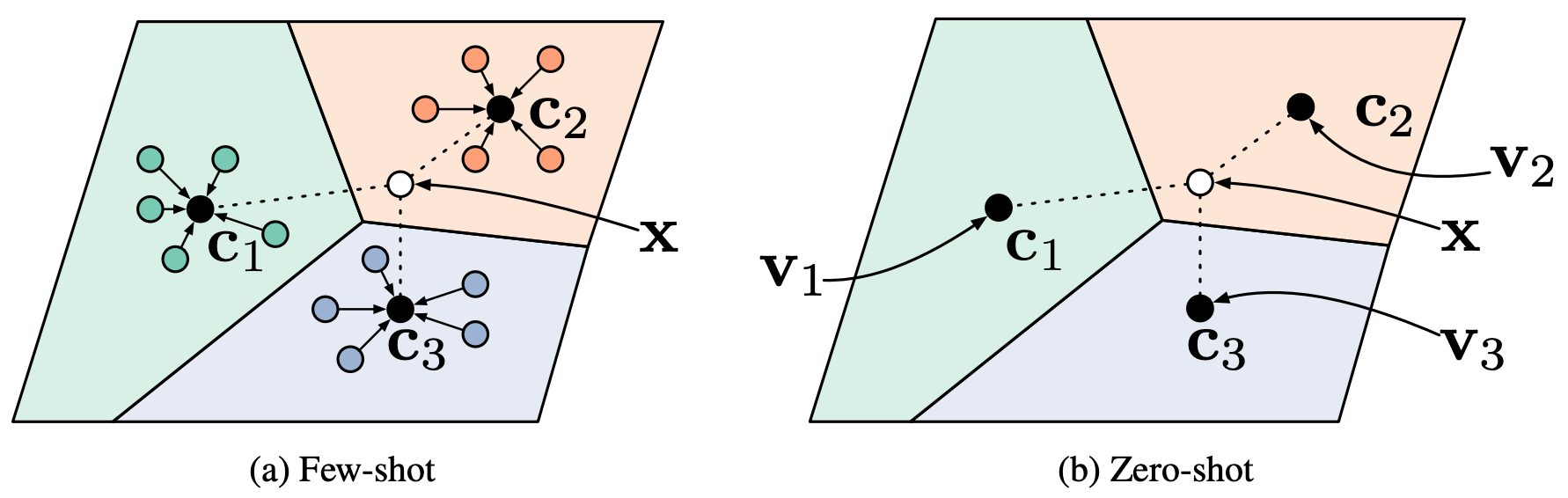}
    \caption{Prototypical network \cite{snell2017prototypical}. In a prototypical network, each sample is encoded into a vector by an encoder. Given a few examples per class (figure a, color dots), a prototype (black dots or $c_1, c_2$, and $c_3$) of a class is computed as the average of all encoded vectors corresponding to that class. Once there is a query $x$, a score distribution is computed based on either the similarity scores or the distances between the given query $x$ and the prototypes. The predicted class is selected to maximize the estimated similarity scores or minimize the estimated distances.}
    \label{fig:prototypical-network}
\end{figure*}

Most of the studies following token classification setting \cite{bronstein-etal-2015-seed,peng-etal-2016-event,lai-nguyen-2019-extending,deng2020meta,lai-etal-2020-extensively} are based on a prototypical network \cite{snell2017prototypical}, which employs a general-purpose event encoder for embed event candidates while the predictions are done using a non-parameterized metric-based classifier. Figure \ref{fig:prototypical-network} demonstrates the idea of a prototypical network. Since the classifiers are non-parametric, these studies mainly explore the methods to improve the event encoder.

\begin{figure}[t]
    \centering
    \includegraphics[width=\linewidth]{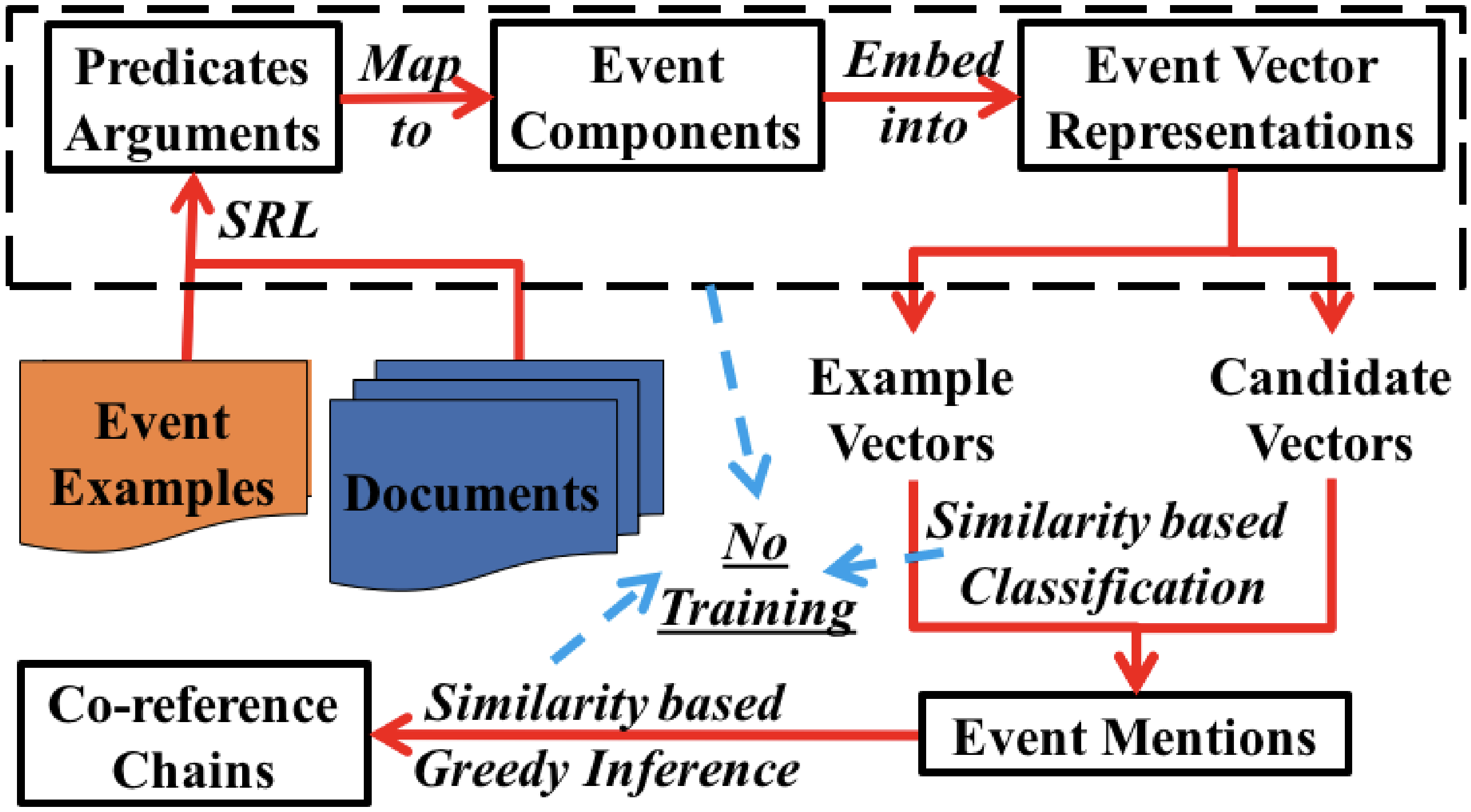}
    \caption{Event Detection framework by \citet{peng-etal-2016-event}}
    \label{fig:peng-etal-2016-event}
\end{figure}

\citet{bronstein-etal-2015-seed} were among the first working in few-shot event detection. They proposed a different training/evaluation for event detection with minimal supervision. They proposed an alternative method, which uses the trigger terms included in the annotation guidelines as seeds for each event type. The model consists of an encoder and a classifier. The encoder embeds a trigger candidate into a fix-size embedding vector. The classifier is an event-independent similarity-based classifier. In this paper, the authors argue that they can eliminate the costly manual annotation for new event types, while the non-parametric classifier does not require a large amount to be trained, in fact, just a few example events at the beginning.
\citet{peng-etal-2016-event} addressed the manual annotation by proposing an event detection and coreference system that requires minimal supervision, in particular, a few training examples. Their approach was built on a key assumption: the semantics of two tasks (i) identifying events that are closely related to some event types and (ii) event coreference are similar. As such, instead of annotating a large dataset for event detection, reformulating the task into semantic similarity can help the model to be trained on a large available corpus of event coreference. As a result, the required data for any new event type is as small as the number of samples in the annotation guidelines. To do that, they use a general purpose nominal and verbial semantic role labeling (SRL) representation to represent the structure of an event. The representation involves multiple semantic spaces, including contextual, topical, and syntactic levels. Figure \ref{fig:peng-etal-2016-event} demonstrates their method in detail.
Similarly, \citet{lai-nguyen-2019-extending} proposed a novel formulation for event detection, namely learning from keywords (LFK) in which each type is described via a few event triggers. They are pre-selected from a pool of known events. In order to encode the sentence, the model contains a CNN-based encoder and a conditional feature-wise attention mechanism to selectively enhance informative features.

\citet{lai-etal-2020-extensively}, \citet{deng2020meta} and \citet{lai-etal-2021-learning} employed the core architecture of the prototypical network while proposed an auxiliary training loss factors during the training process. \citet{lai-etal-2020-extensively} enforce the distances between clusters of samples, namely intra-cluster loss and inter-cluster loss. The intra-cluster loss minimizes the distances between samples of the same class, while the inter-cluster loss maximizes the distances between the prototype of a class and the examples of the other classes. The model also introduces contextualized embedding, which leads to significant performance improvement over ANN or CNN-based encoders. 
\cite{deng2020meta}, on the other hand, proposed a Dynamic-Memory-Based Prototypical Network (DMB-PN). The model uses a Dynamic Memory Network(DMN) to learn better prototypes and produce better event mention encodings. The prototypes are not computed by averaging the supporting events just once, but they are induced from the supporting events multiple times through DMN's multihop mechanism. 
\citet{lai-etal-2021-learning} addressed the outlier and sampling bias in the training process of few-shot event detection. Particularly, in event detection, a null class is introduced to represent samples that are out of the interested classes. These may contain non-interested eventive samples as well as non-eventive samples. As such, this class may inject outlier examples into the support set. As such, they proposed a novel model for the relation between two training tasks in an episodic training setting by allowing interactions between prototypes of two tasks. They also proposed prediction consistency between two tasks so that the trained model would be more resistant to outliers.

\citet{chen-etal-2021-honey} addressed the trigger curse problem in FSL-ED. Particularly, both overfitting and underfitting trigger identification are harmful to the generalization ability or the detection performance of the model, respectively. They argue that the trigger is the confounder of the context and the result of an event. As such, previous models, which are trigger-centric, can easily overfit triggers. To alleviate the trigger overfitting, they proposed a method to intervene in the context by backdoor adjustment during training. 

\begin{figure*}
    \centering
    \includegraphics[width=0.8\textwidth]{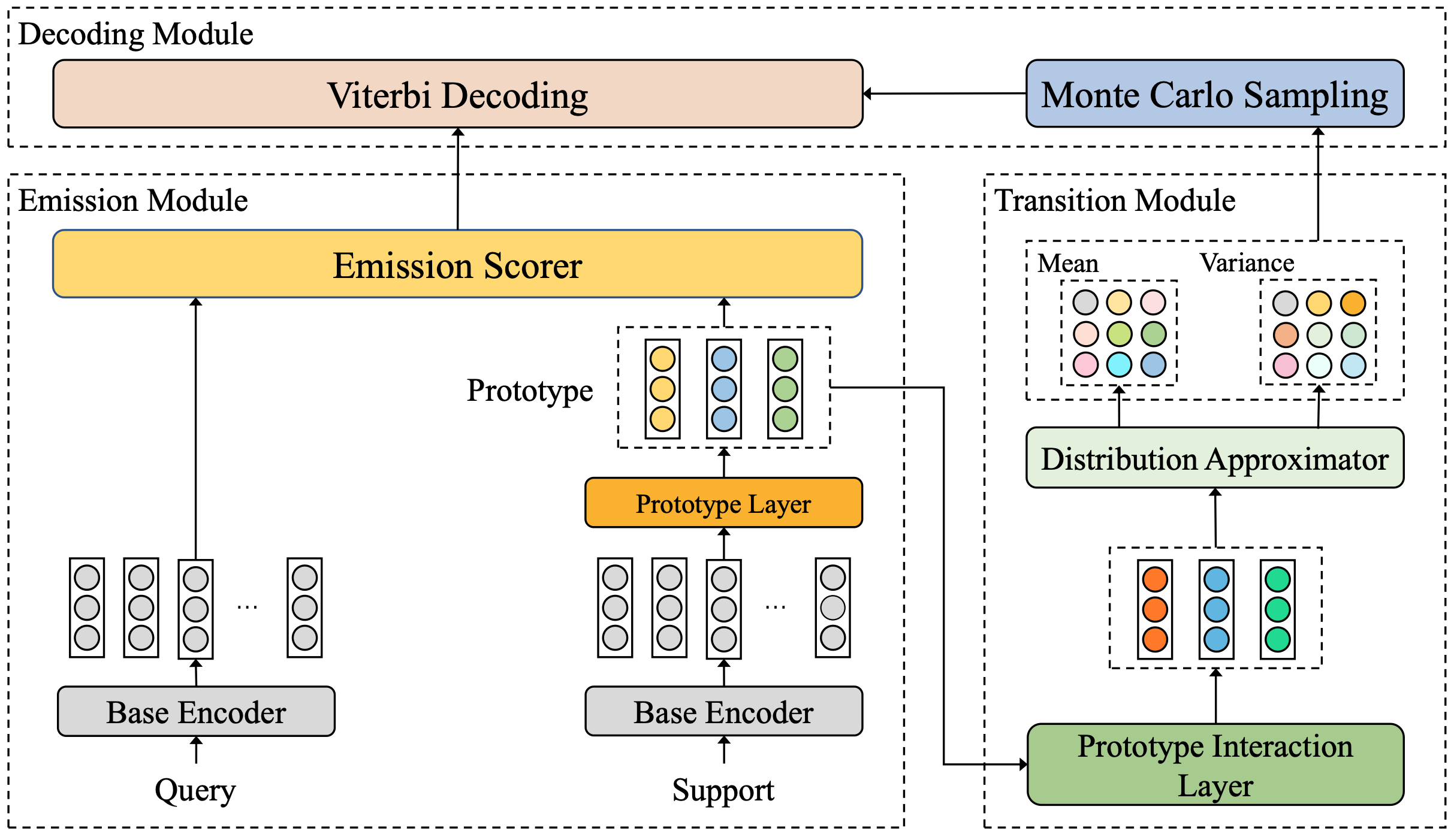}
    \caption{Architecture of the PA-CRF model \cite{cong-etal-2021-shot}. The model consists of 3 modules. The Emission Module computes the emission scores for the query instances based on the prototypes of the support set. The Transition Module estimates the Gaussian distributed transition scores from the prototypes. The Decoding Module consumes the emissions scores and the approximated transition scores to decode the predicted sequences with Monte Carlo sampling.}
    \label{fig:cong21fewshot}
\end{figure*}

Recent work by \citet{shen-etal-2021-adaptive} tackles the low sample diversity in FSL-ED. Their model, Adaptive Knowledge-Enhanced Bayesian Meta-Learning (AKE-BML), introduces external event knowledge as a prior of the event type. To do that, first, they heuristically align the event types in the support set and FrameNet. Then they encode the samples and the aligned examples in the same semantic space using a neural-based encoder. After that, they realign the knowledge representation by using a learnable offset, resulting in a prior knowledge distribution for event types. Then they can generate a posterior distribution for event types. Finally, to predict the label for a query instance, they use the posterior
distribution for prototype representations to classify query instances into event types.

The second FSL-ED setting is based on sequence labeling. The few-shot sequence labeling setting, in general, has been widely studied in named entities recognition \cite{fritzler2019few}. Similarly, \citet{cong-etal-2021-shot} formulated the FSL-ED as a few-shot sequence labeling problem, in which detects the spans of the events and the label of the event at the same time. They argue that previous studies that solve this problem in the \textbf{identify-then-classify} manner suffer from error propagation due to ignoring the discrepancy of triggers between event types. They proposed a CRF-based model called Prototypical Amortized Conditional Random Field (PA-CRF). In order to model the CRF-based classifiers, it is important to approximate the transition scores and emission scores from just a few examples. Their model approximates the transition scores between labels based on the label prototypes. In the meantime, they introduced a Gaussian distribution into the transition scores to alleviate the uncertain estimation of the emission scorer. Figure \ref{fig:cong21fewshot} presents the overview architecture of the model.

\subsection{Cross-lingual}
\label{sec:cross-lingual}

Early studies of cross-lingual event extraction (CLEE) relies on training a statistical model on parallel data for event extraction \cite{chen-ji-2009-one,piskorski-etal-2011-exploring,hsi-etal-2016-leveraging}. Recent methods focus on transferring universal structures across languages \cite{subburathinam-etal-2019-cross,liu-etal-2019-neural,lu-etal-2020-cross,nguyen-nguyen-2021-improving}. There are a few other methods were also studied such as topic modeling \cite{li2011exploiting}, multilingual embedding \cite{mhamdi-etal-2019-contextualized}, and annotation projection \cite{li-etal-2016-learning,lou2022translation}.

Cross-lingual event extraction depends on a parallel corpus for both training and evaluation. However, parallel corpora for this area are scarce. Most of the work in CLEE were done using ACE-2005 \cite{ldc2005ace}, TAC-KBP \cite{Mitamura:15,mitamura2017events}, and TempEval-2 \cite{verhagen-etal-2010-semeval}. These multilingual datasets cover a few popular languages, such as English, Chinese, Arabic, and Spanish. Recently, datasets that cover less common languages, e.g., Polish, Danish, Turkish, Hindi, Urdu, Korean and Japanese, were created for event detection \cite{pouran-ben-veyseh-etal-2022-minion} and event relation extraction \cite{lai-etal-2022-meci}. 

\begin{figure*}[!h]
    \centering
    \includegraphics[width=\textwidth]{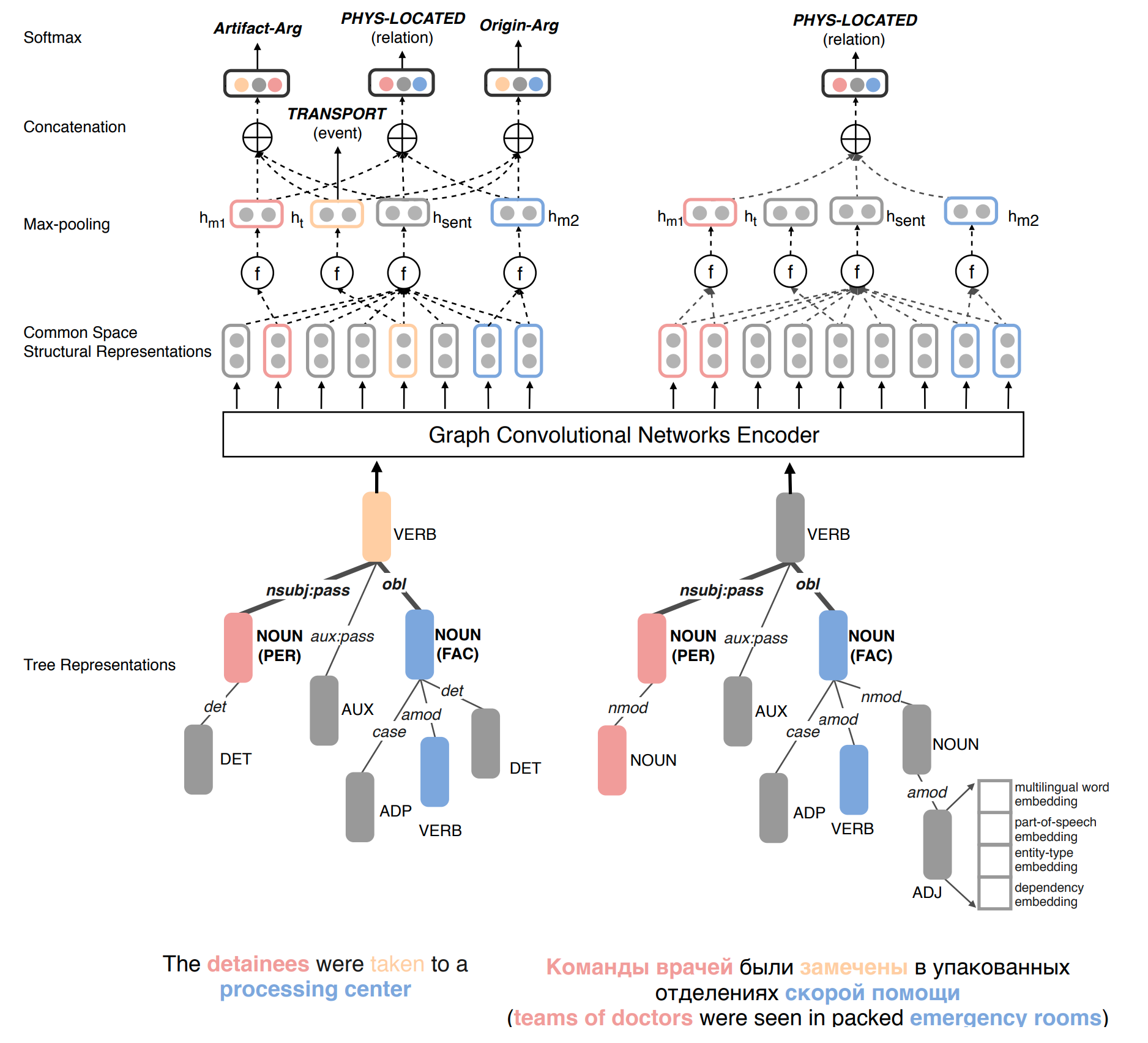}
    \caption{Shared structures of different event mentions, written in English and Russian, of the same event type \cite{subburathinam-etal-2019-cross}.}
    \label{fig:subburathinam19crosslingual}
\end{figure*}

Due to data scarcity in target languages, the model trained on limited data might not be able to predict a wide range of events. Therefore, generating more data from the existing corpus in the source language is a trivial method. \citet{li-etal-2016-learning} proposed a projection algorithm to mine shared hidden phrases and structures between two languages (i.e., English and Chinese). To do that, they project seed phrases back and forth multiple rounds between the two languages using parallel corpora to obtain a diverse set of closely related phrases. The captured phrases are then used to train an ED model. This method was shown to effectively improve the diversity of the recognized events. \citet{lou2022translation} addressed the problem of noise appearing in the translated corpus. They proposed an annotation projection approach that combines the translation projection and the event argument extraction task training step to alleviate the additional noise through implicit annotation projection. First, they translate the source language corpus into the target language using a multilingual machine translation model. To reduce the noise of the translated data, instead of training the model directly from them, they use a multilingual embedding to embed the source language data and the translated derivatives in the target language into the same vector space. Their representations are then aligned using optimal transport. They proposed two additional training signals that either reduce the alignment scores or the prediction based on the aligned representation. \citet{phung-etal-2021-learning} explored the cross-lingual transfer learning for event coreference resolution task. They introduced the language adversarial neural network to help the model distinguish texts from the source and target languages. This helps the model improve the generalization over languages for the task. Similar to \cite{lou2022translation}, the work by \citet{phung-etal-2021-learning} introduced an alignment method based on multiple views of the text from the source and the target languages. They further introduced optimal transport to better select edge examples in the source and target languages to train the language discriminator.

Multilingual embedding plays an important role in transferring knowledge between languages. There have been many multilingual contextualized embedding built for a large number of languages such as FastText \cite{joulin-etal-2018-loss}, MUSE \cite{lample2017unsupervised}, mBERT \cite{devlin-etal-2019-bert}, mBART \cite{liu-etal-2020-multilingual-denoising}, XLM-RoBERTa \cite{conneau-etal-2020-unsupervised}, and mT5/mT6 \cite{xue-etal-2021-mt5,chi-etal-2021-mt6}. \cite{mhamdi-etal-2019-contextualized} compared FastText, MUSE and mBERT. The results show that multilingual embeddings help transfer knowledge from English data to other languages i.e. Chinese and Arabic. The performance boost is significant when all multilingual are added to train the model. In cross-lingual event extraction, various multilingual embeddings have been employed thanks to their robustness and transferability. However, models that were trained on multilingual embedding still suffers from performance drop in zero-shot cross-lingual setting if the monolingual model is trained on a large enough target dataset and a good enough monolingual contextualized embedding \cite{lai-etal-2022-meci}.

Most of the recent methods for cross-lingual event extraction are done via transferring shared features between languages, such as syntactic structures (e.g., part-of-speech, dependency tree), semantic features (e.g., contextualized embedding), and relation structures (e.g., entity relation). \citet{subburathinam-etal-2019-cross} addressed the suitability of transferring cross-lingual structures for the event and relation extraction tasks. In particular, they exploit relevant language-universal features for relation and events such as symbolic features (i.e., part-of-speech and dependency path) and distributional features (i.e., type representation and contextualized representation) to transfer those structures appearing in the source language corpus to the target language. Figure \ref{fig:subburathinam19crosslingual} presents an example of shared structures of two different sentences in two languages (i.e., English and Russian). Thanks to this similarity, they encode all the entity mentions, event triggers, and event context from both languages into a complex shared cross-lingual vector space using a graph convolutional neural network. Hence, once the model is trained on English, this shared structural knowledge will be transferred to the target languages, such as Russian. \cite{liu-etal-2019-neural} addressed two issues in cross-lingual transfer learning: (i) how to build a lexical mapping between languages and (ii) how to manage the effect of the word-order differences between different languages. First, they employ a context-dependent translation method to construct the lexical mapping between languages by first retrieving k nearest neighbors in a shared vector space, then reranking the candidates using a context-aware selective attention mechanism. To encode sentences with language-dependent word order, a GCN model is employed to encode the sentence. To enrich the features for the cross-lingual event argument extraction model, \citet{nguyen-nguyen-2021-improving} employ three types of connection to build a feature-expanded graph. Particularly, the core of the graph is derived from the dependency graph used in many other studies to capture syntactic features. They introduced two additional connections to capture semantic similarity and the universal dependency relations of the word pairs. Based on the assumption that most concepts are universal across languages, hence, similarities between words and representing concepts are also universal. They employ a multilingual contextualized embedding to obtain the word representation, then compute a similarity score between words in a sentence. Secondly, they argue that the relation types play an important role in how strong the connection is. Therefore, another connection set of weights is computed based on the dependency relation type between two connected words. Finally, the additional edge weights are added to the graph, scaling to the extent of the similarity score of the relation.

\clearpage
\bibliographystyle{acl_natbib}
\bibliography{anthology,ref}

\end{document}